\documentclass{article}

\usepackage[english]{babel}

\usepackage[letterpaper,top=2cm,bottom=2cm,left=3cm,right=3cm,marginparwidth=1.75cm]{geometry}

\usepackage{amsmath}
\usepackage{graphicx}
\usepackage[colorlinks=true, allcolors=blue]{hyperref}
\usepackage{booktabs} 
\usepackage{graphicx}
\usepackage{changepage} 
\usepackage{csquotes}
\usepackage[style=alphabetic,maxnames=1,minnames=1,maxbibnames=3]{biblatex}
\usepackage{xcolor}
\usepackage{colortbl}
\usepackage{pgf}

\def\colornum#1{
\newcommand{\MaxNumber}{100} 
\newcommand{\MinNumber}{30}   
\pgfmathsetmacro{\PercentColor}{max(min(100.0*(#1 - \MinNumber)/(\MaxNumber-\MinNumber),100.0),0.00)}
\hspace{-0.33em}\colorbox{pink!\PercentColor!white}{#1}
}

\title{MTFinEval:A Multi-domain Chinese Financial Benchmark with Eurypalynous questions}
\author{Ke Jin, Xinyu Liu}
\author{ 
Xinyu Liu\textsuperscript{1},
Ke Jin\textsuperscript{1}\\
\textsuperscript{1}Beihang University
}

\begin{document}
\maketitle

\begin{abstract}
With the emergence of more and more economy-specific LLMS, how to measure whether they can be safely invested in production becomes a problem.
Previous research has primarily focused on evaluating the performance of LLMs within specific application scenarios. 
However, these benchmarks cannot reflect the theoretical level and generalization ability, and the backward datasets are increasingly unsuitable for problems in real scenarios.
In this paper, we have compiled a new benchmark, MTFinEval, focusing on the LLMs' basic knowledge of economics, which can always be used as a basis for judgment.
To examine only theoretical knowledge as much as possible, MTFinEval is build with foundational questions from university textbooks,and exam papers in economics and management major.
Aware of the overall performance of LLMs do not depend solely on one subdiscipline of economics, MTFinEval comprise 360 questions refined from six major disciplines of economics, and reflect capabilities more comprehensively.
Experiment result shows all LLMs perform poorly on MTFinEval, which proves that our benchmark built on basic knowledge is very successful.
Our research not only offers guidance for selecting the appropriate LLM for specific use cases, but also put forward increase the rigor reliability of LLMs from the basics.

\end{abstract}

\section{Introduction}
In the realm of economics, LLMs offer economists and policymakers unique insights\parencite{Li2023CFGPTCF}, thereby enhancing the efficiency of economic industry development\parencite{Zha2023TableGPTTU}\parencite{Zhang2024TableLLMET}. 
For example, CatMemo\parencite{Cao2024CatMemoAT} uses LLM for stock trading, FINANCEBENCH\parencite{Islam2023FinanceBenchAN} uses LLM for company earnings analysis, and FinPT\parencite{Yin2023FinPTFR} uses LLM for risk forecasting. 
Financial data grows exponentially in both volume and complexity. 
However, the task-oriented benchmarks\parencite{Lei2023CFBenchmarkCF} \parencite{Araci2019FinBERTFS}, composed of specific past events are gradually deviating from the actual situation. 
No matter a drastic change in policy between countries, or disruptive innovation in new technologies such as AI, it will lead to dramatic changes in economic phenomena. 
Furthermore, the comprehensive capabilities of LLMs in the field of economics cannot be fully assessed by a single task requirement. 
Therefore we create a benchmark, MTFinEval, to examine LLM theoretical knowledge across a wide range of economics fields.

In this article, we narrow focus to the cognitive level of theoretical knowledge within LLMs because theoretical knowledge is the most fundamental requirement. 
It not only shapes the model’s grasp of problems but also forms the basis for task execution.
Our analysis breaks down the comprehensive capabilities of LLMs across numerous sub-aspects. 
This approach aims to identify the reasons behind the subpar performance of LLMs when tackling complex tasks, offering a clearer direction for understanding their limitations.

The benchmark, MTFinEval, covers six fields of management, accounting, e-commerce, strategic management of enterprise, macroeconomics and microeconomics, and covers multiple dimensions such as economic indicators, financial technology, financial law and economic phenomena. 
For all questions, the ability of the model is examined in the form of question and answer, and the correct rate of the model is directly calculated. 
Through this dataset, we aim to assess the multi-faceted capabilities\parencite{Liu2019MultiTaskDN}\parencite{Ma2018ModelingTR}\parencite{Kendall2017MultitaskLU}\parencite{Liu2018EndToEndML} of LLMs in the field of finance, including but not limited to data understanding, logical reasoning, and situational adaptation.

\begin{figure}
    \centering
    \includegraphics[width=0.6\linewidth]{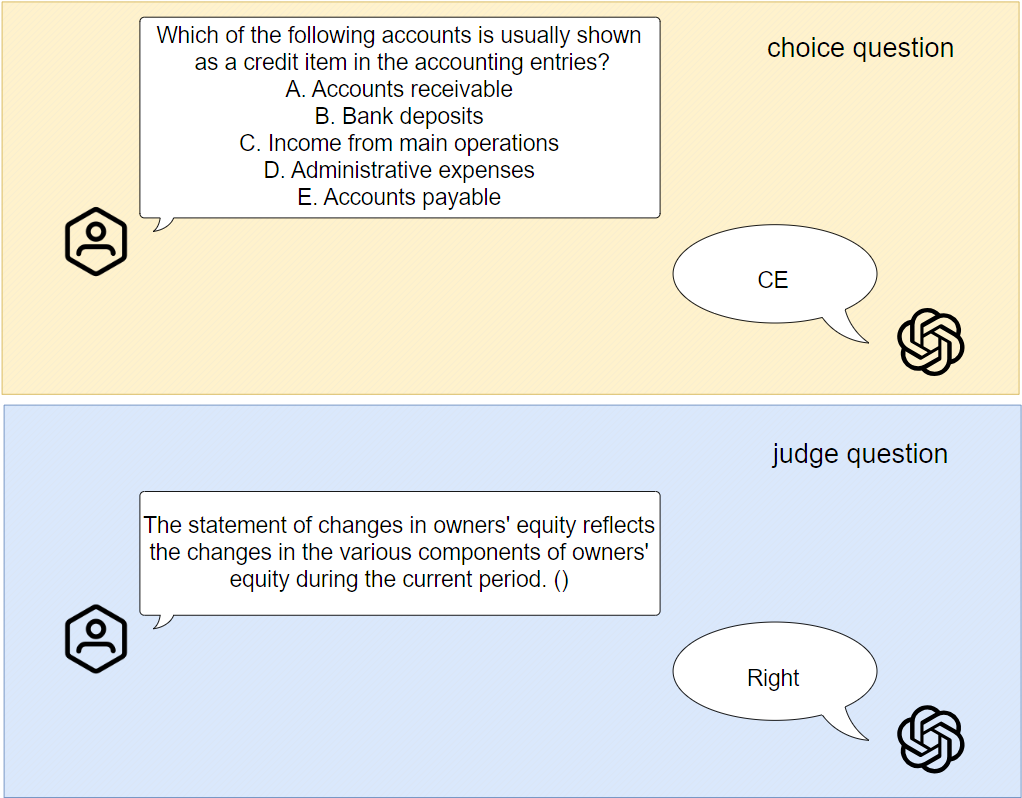}
    \caption{Answer pairs on different question types.}
    \label{fig:enter-label}
\end{figure}

\section{Related Work}
In finance and economics, the application of Large Language Models (LLMs) is emerging as an essential tool for in-depth market analysis\parencite{Zhang2023XuanYuan2A}\parencite{Cornucopia-LLaMA-Fin-Chinese}\parencite{paeg}, precise investment advice, and effective risk assessment\parencite{ganlm}.
This paper aims to delve into the specialized development and potential of LLMs within the financial domain, highlighting the significance of systematic evaluation and theoretical integration.

Firstly, BloombergGPT\parencite{Wu2023BloombergGPTAL}, an LLM tailored for the financial sector, exhibits significant potential in performing financial natural language processing (NLP) tasks.
Its capability to manage complex financial data and tasks, such as analyzing market trends and generating investment reports\parencite{Koa2024LearningTG}, illustrates the promise of LLM applications in finance.
Similarly, FinMA, through command tuning, adeptly handles a variety of financial NLP tasks\parencite{Sinha2020ImpactON}\parencite{Zhou2021TradeTE}\parencite{Maia2018WWW18OC}\parencite{CIKM2020MAEC} including sentiment analysis, event detection\parencite{cortis-etal-2017-semeval}, and risk assessment\parencite{Alvarado2015DomainAO}\parencite{10386611}, further demonstrating the broad application potential of LLMs in the financial sector.

Nonetheless, systematic evaluation of LLMs' performance in finance is crucial.
The introduction of the EconNLI dataset\parencite{Guo2024EconNLIEL}, designed to assess LLMs' knowledge and competence in economic reasoning, exposes potential deficiencies in LLMs' economic reasoning abilities\parencite{Park2023MachineLB}.
This underscores the necessity for thorough evaluation of these models.
Additionally, models that incorporate economic theory excel in financial analysis and forecasting, emphasizing the importance of integrating economic knowledge when developing financial LLMs.
Evaluating LLMs’ performance in specialized areas is crucial for gauging their true capabilities.
While existing assessments often concentrate on general NLP tasks relevant to finance, such as causal reasoning, text classification, and predictive analytics, there is a dearth of systematic evaluations tailored to the financial and economic sectors.
It becomes evident that a deep grasp of economic principles significantly enhances LLMs’ financial task performance\parencite{Zhai2024ActionsSL}.

The application of multimodal learning in financial forecasting also presents new opportunities.
For instance, MONOPOLY\parencite{Mathur2022MONOPOLYFP} leverages multimodal cues to forecast finances from monetary policy meeting videos, offering a novel perspective on market dynamics understanding.
Concurrently, the exploration of cross-language and zero-sample learning capabilities\parencite{Jin2024ZeroShotCR}, such as the zero-sample cross-language named entity recognition method introduced by CROP\parencite{Yang2022CROPZC}\parencite{Yang2020AlternatingLM}, opens up new possibilities for multilingual\parencite{um4} financial applications and aids in processing financial documents in various languages.

In conclusion, LLMs in the financial and economic fields demonstrate unique value and potential.
To enhance the performance and reliability of these models\parencite{Huang2022MBCTTF}, a deeper understanding and integration of economic theory\parencite{10.1007/978-3-030-45439-5_46}\parencite{Elman1993LearningAD}\parencite{Fan2018LearningTT} are required.

\section{MTFinEval Benchmark}
\subsection{Dataset Statistics}
Our research is dedicated to strengthening the capacity of these models to tackle economic challenges fundamentally and systematically. 

The MTFinEval dataset comprises 360 university economics questions, spanning six major sub-topics: macroeconomics, microeconomics, accounting, management, e-commerce, and strategic—management. 
Each subtopic includes single choice, multiple choice, and true or false questions.
All questions and answers are manually extracted from college textbooks\parencite{Li2023TextbooksAA} and exam papers to ensure they are foundational and introductory.
The data  collection process has been meticulously scrutinized through a series of systematic checks.
We began by manually entering the paper questions into CSV files, ensuring the completeness of each entry.  

Subsequently, to verify the accuracy of the questions and answers, we enlisted the expertise of six economists, 
each with high proficiency in various subfields.
They were tasked with reviewing the clarity of the descriptions and the correctness of the answers, for which they were compensated at a rate of \$3 per question.
In cases where a question was deemed questionable during the expert review stage, a collective discussion was held among all experts to determine the correct answer and to decide if the question should be omitted. The types and numbers of questions for all subtopics are shown in Figure 1.
\begin{figure}
    \centering
    \includegraphics[width=0.4\linewidth]{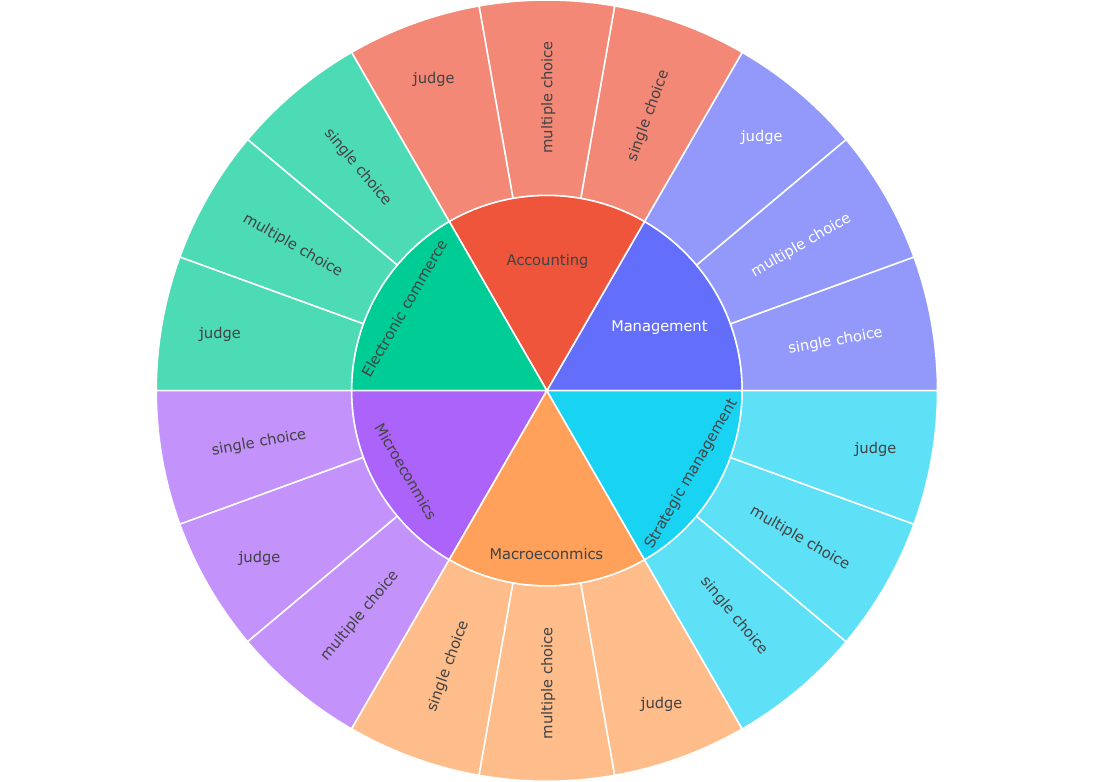}
    \caption{types and numbers of questions}
    \label{types and numbers of questions}
\end{figure}

\subsection{Formulation}

For problem type $T$, given the large model $\mathcal{M}$, the corpus set $d$ used for training and fine-tuning the large model, input the question q, and the large model returns the answer $a$\parencite{Narayan2018DontGM}.

In the zero-shot scenario for true or false statements, the objective function is defined as follows:\\
\begin{MiddleEquation}
\begin{align}
    r = \underset{a \in \{0, 1\}}{\arg\max} P(a|q^T; \mathcal{M}, d^T) \label{equ:single-choice_loss}
\end{align}
\end{MiddleEquation}
Similarly, for single-choice questions in the zero-shot scenario, the objective function is specified as:
\begin{MiddleEquation}
\begin{align}
    r = \underset{a \in \{A, B, C, D\}}{\arg\max} P(a|q^T; \mathcal{M}, d^T) \label{equ:multiple-choice_loss}
\end{align}
\end{MiddleEquation}
Lastly, for multiple-choice questions in the zero-shot scenario, the objective function is:
\begin{MiddleEquation}
\begin{align}
   r = \underset{a \in \{A, B, C, AB, AC, BC, ABC ...\}}{\arg\max} P(a|q^T; \mathcal{M}, d^T) \label{equ:judge_loss}
\end{align}
\end{MiddleEquation}

On the surface, choices and judgments are both classify problems.
Fundamentally, for the decoder only model, 
the underlying task is a generation task. 
In the training of LLMs, While pursuing to reduce cross entropy, maximum likelihood estimation is also used to reproduce similar sentences in the corpus.
Therefore, although only the selection of the judgment problem, the bottom layer still recalls the relevant corpus content during the training of the LLMs.

\section{Experiments}

\subsection{Experimental Setup}

For problem type T, given the large model M,the corpus set d used for training and finetuning the large model, input the question q, and the large model returns the answer a. In the zero-shot scenario for judge questions the objective function is define To simulate the most direct use cases effectively, we introduce a brief prompt to the question. This prompt includes a description of the question type, all under zero-shot conditions\parencite{Brown2020LanguageMA}.
The prompts for the three question types are similiar with follows\parencite{Wei2022ChainOT}\parencite{xcot}\parencite{low_resource_template}\parencite{hlt_mt}\parencite{Tai2023ExploringCS}: multiple choice: You are a financial knowledge expert, please combine your financial knowledge to answer the following multiple choice, note that there is not only one answer to multiple choice, to think and reason whether each option is correct but do not  
output thinking process, directly return the option letter in markdown format, each option in the answer is separated by a space.
When it comes to judgment questions, the LLM is instructed to provide answers limited to ”yes,” ”wrong,” or ”do not know.”.
Subsequently, key words from these answers are extracted and mapped onto these three specific options.
In the realm of traditional economics, LLMs are typically based on the Llama series\parencite{Touvron2023Llama2O}\parencite{Dubey2024TheL3} for processing.
To promote decentralization, the models participating in this evaluation are exclusively open-source models that were released in the year 2024.
\subsection{Results}
\begin{table}[]
\scalebox{0.6}{
\begin{tabular}{cccccccc}
\toprule 
model & macroeconomics& microeconomics & accounting & management & e-commerce & strategic-management & comprehensive\\
\midrule 
Qwen/Qwen2-7B-Instruct & \colornum{55.0} & \colornum{66.67} & \colornum{80.0} & \colornum{51.67} & \colornum{65.0} & \colornum{61.67} & \colornum{63.33} \\
Qwen/Qwen2-1.5B-Instruct & \colornum{23.33} & \colornum{30.0} & \colornum{50.0} & \colornum{40.0} & \colornum{50.0} & \colornum{46.67} & \colornum{40.0} \\
Qwen/Qwen1.5-7B-Chat& \colornum{40.0} & \colornum{50.0} & \colornum{66.67} & \colornum{45.0} & \colornum{60.0} & \colornum{51.67} & \colornum{52.22} \\
THUDM/glm-4-9b-chat& \colornum{53.33} & \colornum{55.0} & \colornum{70.0} & \colornum{46.67} & \colornum{66.67} & \colornum{56.67} & \colornum{58.06} \\
THUDM/chatglm3-6b& \colornum{26.67} & \colornum{35.0} & \colornum{35.0} & \colornum{30.0} & \colornum{50.0} & \colornum{35.0} & \colornum{35.28} \\
01-ai/Yi-1.5-9B-Chat-16K & \colornum{48.33} & \colornum{55.0} & \colornum{80.0} & \colornum{48.33} & \colornum{60.0} & \colornum{56.67} & \colornum{58.06} \\
01-ai/Yi-1.5-6B-Chat & \colornum{40.0} & \colornum{53.33} & \colornum{58.33} & \colornum{48.33} & \colornum{68.33} & \colornum{53.33} & \colornum{53.61} \\
google/gemma-2-9b-it & \colornum{56.67} & \colornum{50.0} & \colornum{56.67} & \colornum{50.0} & \colornum{58.33} & \colornum{65.0} & \colornum{56.11} \\
meta-llama/Meta-Llama-3-8B-Instruct & \colornum{30.0} & \colornum{33.33} & \colornum{45.0} & \colornum{38.33} & \colornum{38.33} & \colornum{55.0} & \colornum{40.0} \\
\bottomrule
\end{tabular}
}
\caption{accuracy(\%) of different models on different sub-disciplines}
\end{table}

Regardless of little changes in the LLMs' architecture, the diverse answers provided by various LLMs can reveal the extent and quality of the training data used. This insight can then be used to reasonably gauge the model's capability level.
Specialization in Subjects: Certain models demonstrate expertise in specific domains. For instance, the 01-ai/Yi-1.5-9B-Chat-16K\parencite{Young2024YiOF} model excels in accounting, e-commerce, and strategic management, suggesting a rich dataset in these areas. This proficiency likely stems from its capacity to interpret complex financial data, manage online platforms efficiently, and formulate strategic business plans.
Macroeconomic Understanding: The THUDM/glm-4-9b-chat\parencite{Zeng2024ChatGLMAF}\parencite{Bommasani2021OnTO} model's elevated score in macroeconomics indicates a robust grasp of economic trends and policy impacts. This proficiency may be due to its exposure to a diverse range of economic data, enabling precise assessments.
Strategic Management Insight: The google/Gemma-2-9B-IT model's high strategic management score reflects its potential to support business planning. Its ability to integrate information from various sources to offer valuable strategic insights is likely the reason for this performance.
Overall Performance: The Qwen/Qwen2-7B-Instruct\parencite{Bai2023QwenTR}\parencite{Yang2024Qwen2TR} model's lead in overall performance with a 63.33 score signifies its emergence as a leading contender in the field of financial language model technology. Its comprehensive score not only reflects its proficiency in individual subjects but also suggests a robust and well-rounded training approach that has equipped it to handle a variety of economic analyses. This model's performance is indicative of its potential to become the new baseline, or pedestal\parencite{Nori2023CanGF}, for financial LLMs\parencite{Chen2023TigerBotAO}, possibly outperforming or replacing previous models such as Llama\parencite{Cui2023EfficientAE} in certain applications.
Reflection on Weaker Performances: Models like the meta-llama/Meta-Llama-3-8B-Instruct, which scored lower overall, may have been trained on less diverse or pertinent data for certain subjects, or they may have architectural limitations hindering effective processing and analysis of economic information.
Implications for Model Improvement: To enhance performance in specific subjects, models should be trained with more domain-specific data. Moreover, ongoing updates and refinements to the model architecture, informed by performance data, can further augment capabilities.
In conclusion, LLM performance in economic subjects mirrors the quality of their training data and architecture. While specialized models are better suited for particular applications, a well-rounded model like Qwen/Qwen2-7B-Instruct serves as a multifaceted tool for a range of economic analyses. Ongoing refinement of training data and model architecture is essential for advancing their capabilities in economics.

\section{Conclusion}
\begin{figure}
    \centering
    \includegraphics[width=0.8\linewidth]{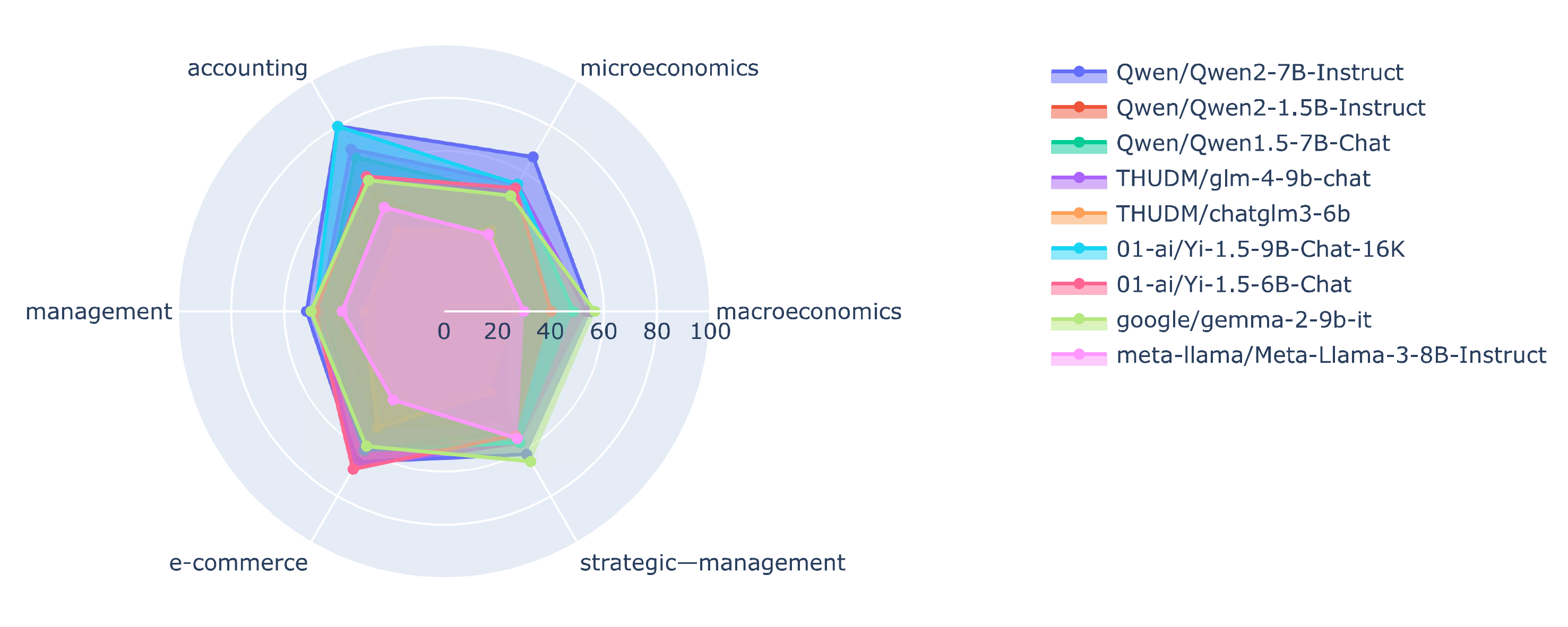}
    \caption{subject and type radars}
    \label{fig:enter-label}
\end{figure}
In this study, we introduced MTFinEval, a comprehensive multi-domain benchmark, consisting of 360 questions across six major economic disciplines. It provides a rigorous assessment tool to evaluate the fundamental economic knowledge of LLMs, for the purpose of ensuring that the LLM still makes the right analysis in a rapidly changing environment. The experimental results demonstrated that current LLMs, generally perform poorly on this benchmark, highlighting significant gaps in their theoretical understanding of economics. This research contributes to the field by offering a robust evaluation framework that can guide future improvements in LLMs, ensuring they are better equipped for complex and dynamic economic environments.

\printbibliography

@article{Wu2023BloombergGPTAL,
  title={BloombergGPT: A Large Language Model for Finance},
  author={Shijie Wu and Ozan Irsoy and Steven Lu and Vadim Dabravolski and Mark Dredze and Sebastian Gehrmann and Prabhanjan Kambadur and David Rosenberg and Gideon Mann},
  journal={ArXiv},
  year={2023},
  volume={abs/2303.17564},
  url={https://api.semanticscholar.org/CorpusID:257833842}
}

@article{Koa2024LearningTG,
  title={Learning to Generate Explainable Stock Predictions using Self-Reflective Large Language Models},
  author={Kelvin J.L. Koa and Yunshan Ma and Ritchie Ng and Tat-Seng Chua},
  journal={Proceedings of the ACM on Web Conference 2024},
  year={2024},
  url={https://api.semanticscholar.org/CorpusID:267500314}
}

@article{Chen2023TigerBotAO,
  title={TigerBot: An Open Multilingual Multitask LLM},
  author={Ye Chen and Wei Cai and Liangming Wu and Xiaowei Li and Zhanxuan Xin and Cong Fu},
  journal={ArXiv},
  year={2023},
  volume={abs/2312.08688},
  url={https://api.semanticscholar.org/CorpusID:266209771}
}

@inproceedings{Cao2024CatMemoAT,
  title={CatMemo at the FinLLM Challenge Task: Fine-Tuning Large Language Models using Data Fusion in Financial Applications},
  author={Yupeng Cao and Zhiyuan Yao and Zhi Chen and Zhiyang Deng},
  year={2024},
  url={https://api.semanticscholar.org/CorpusID:270878423}
}

@article{Yin2023FinPTFR,
  title={FinPT: Financial Risk Prediction with Profile Tuning on Pretrained Foundation Models},
  author={Yuwei Yin and Yazheng Yang and Jian Yang and Jian Yang and Qi Liu},
  journal={ArXiv},
  year={2023},
  volume={abs/2308.00065},
  url={https://api.semanticscholar.org/CorpusID:260350873}
}

@article{Islam2023FinanceBenchAN,
  title={FinanceBench: A New Benchmark for Financial Question Answering},
  author={Pranab Islam and Anand Kannappan and Douwe Kiela and Rebecca Qian and Nino Scherrer and Bertie Vidgen},
  journal={ArXiv},
  year={2023},
  volume={abs/2311.11944},
  url={https://api.semanticscholar.org/CorpusID:265294665}
}

@article{Zhang2023XuanYuan2A,
  title={XuanYuan 2.0: A Large Chinese Financial Chat Model with Hundreds of Billions Parameters},
  author={Xuanyu Zhang and Qing Yang and Dongliang Xu},
  journal={Proceedings of the 32nd ACM International Conference on Information and Knowledge Management},
  year={2023},
  url={https://api.semanticscholar.org/CorpusID:258833440}
}

@inproceedings{Guo2024EconNLIEL,
  title={EconNLI: Evaluating Large Language Models on Economics Reasoning},
  author={Yue Guo and Yi Yang},
  year={2024},
  url={https://api.semanticscholar.org/CorpusID:270870818}
}

@misc{Cornucopia-LLaMA-Fin-Chinese,
  title={Cornucopia-LLaMA-Fin-Chinese},
  author={YangMu Yu},
  year={2023},
  publisher = {GitHub},
  journal = {GitHub repository},
  howpublished = {\url{https://github.com/jerry1993-tech/Cornucopia-LLaMA-Fin-Chinese}},
}

@article{Bai2023QwenTR,
  title={Qwen Technical Report},
  author={Jinze Bai and Shuai Bai and Yunfei Chu and Zeyu Cui and Kai Dang and Xiaodong Deng and Yang Fan and Wenhang Ge and Yu Han and Fei Huang and Binyuan Hui and Luo Ji and Mei Li and Junyang Lin and Runji Lin and Dayiheng Liu and Gao Liu and Chengqiang Lu and K. Lu and Jianxin Ma and Rui Men and Xingzhang Ren and Xuancheng Ren and Chuanqi Tan and Sinan Tan and Jianhong Tu and Peng Wang and Shijie Wang and Wei Wang and Shengguang Wu and Benfeng Xu and Jin Xu and An Yang and Hao Yang and Jian Yang and Jian Yang and Shusheng Yang and Yang Yao and Bowen Yu and Yu Bowen and Hongyi Yuan and Zheng Yuan and Jianwei Zhang and Xing Zhang and Yichang Zhang and Zhenru Zhang and Chang Zhou and Jingren Zhou and Xiaohuan Zhou and Tianhang Zhu},
  journal={ArXiv},
  year={2023},
  volume={abs/2309.16609},
  url={https://api.semanticscholar.org/CorpusID:263134555}
}

@article{xcot,
  author       = {Linzheng Chai and
                  Jian Yang and
                  Tao Sun and
                  Hongcheng Guo and
                  Jiaheng Liu and
                  Bing Wang and
                  Xinnian Liang and
                  Jiaqi Bai and
                  Tongliang Li and
                  Qiyao Peng and
                  Zhoujun Li},
  title        = {xCoT: Cross-lingual Instruction Tuning for Cross-lingual Chain-of-Thought
                  Reasoning},
  journal      = {arXiv preprint arXiv:2401.07037},
  volume       = {abs/2401.07037},
  year         = {2024},
  url          = {https://doi.org/10.48550/arXiv.2401.07037},
  doi          = {10.48550/ARXIV.2401.07037},
  eprinttype    = {arXiv},
  eprint       = {2401.07037},
}

@inproceedings{low_resource_template,
  author       = {Ze Yang and
                  Wei Wu and
                  Jian Yang and
                  Can Xu and
                  Zhoujun Li},
  editor       = {Kentaro Inui and
                  Jing Jiang and
                  Vincent Ng and
                  Xiaojun Wan},
  title        = {Low-Resource Response Generation with Template Prior},
  booktitle    = {Proceedings of the 2019 Conference on Empirical Methods in Natural
                  Language Processing and the 9th International Joint Conference on
                  Natural Language Processing, {EMNLP-IJCNLP} 2019, Hong Kong, China,
                  November 3-7, 2019},
  pages        = {1886--1897},
  publisher    = {Association for Computational Linguistics},
  year         = {2019},
  url          = {https://doi.org/10.18653/v1/D19-1197},
  doi          = {10.18653/V1/D19-1197},
  bibsource    = {dblp computer science bibliography, https://dblp.org}
}

@inproceedings{um4,
  author       = {Jian Yang and
                  Yuwei Yin and
                  Shuming Ma and
                  Dongdong Zhang and
                  Shuangzhi Wu and
                  Hongcheng Guo and
                  Zhoujun Li and
                  Furu Wei},
  editor       = {Luc De Raedt},
  title        = {{UM4:} Unified Multilingual Multiple Teacher-Student Model for Zero-Resource
                  Neural Machine Translation},
  booktitle    = {Proceedings of the Thirty-First International Joint Conference on
                  Artificial Intelligence, {IJCAI} 2022, Vienna, Austria, 23-29 July
                  2022},
  pages        = {4454--4460},
  publisher    = {ijcai.org},
  year         = {2022},
  url          = {https://doi.org/10.24963/ijcai.2022/618},
  doi          = {10.24963/IJCAI.2022/618},
  bibsource    = {dblp computer science bibliography, https://dblp.org}
}

@inproceedings{hlt_mt,
  author       = {Jian Yang and
                  Yuwei Yin and
                  Shuming Ma and
                  Dongdong Zhang and
                  Zhoujun Li and
                  Furu Wei},
  editor       = {Luc De Raedt},
  title        = {High-resource Language-specific Training for Multilingual Neural Machine
                  Translation},
  booktitle    = {Proceedings of the Thirty-First International Joint Conference on
                  Artificial Intelligence, {IJCAI} 2022, Vienna, Austria, 23-29 July
                  2022},
  pages        = {4461--4467},
  publisher    = {ijcai.org},
  year         = {2022},
  url          = {https://doi.org/10.24963/ijcai.2022/619},
  doi          = {10.24963/IJCAI.2022/619},
  bibsource    = {dblp computer science bibliography, https://dblp.org}
}

@article{Tai2023ExploringCS,
  title={Exploring Chain-of-Thought Style Prompting for Text-to-SQL},
  author={Chang-You Tai and Ziru Chen and Tianshu Zhang and Xiang Deng and Huan Sun},
  journal={ArXiv},
  year={2023},
  volume={abs/2305.14215},
  url={https://api.semanticscholar.org/CorpusID:258841412}
}

@article{Zha2023TableGPTTU,
  title={TableGPT: Towards Unifying Tables, Nature Language and Commands into One GPT},
  author={Liangyu Zha and Junlin Zhou and Liyao Li and Rui Wang and Qingyi Huang and Saisai Yang and Jing Yuan and Changbao Su and Xiang Li and Aofeng Su and Zhang Tao and Chengcheng Zhou and Kaizhe Shou and Miao Wang and Wufang Zhu and Guoshan Lu and Chaonan Ye and Yali Ye and Wen-song Ye and Yiming Zhang and Xing-yan Deng and J. Xu and Haobo Wang and Gang Chen and Junbo Jake Zhao},
  journal={ArXiv},
  year={2023},
  volume={abs/2307.08674},
  url={https://api.semanticscholar.org/CorpusID:259937503}
}

@article{Zhang2024TableLLMET,
  title={TableLLM: Enabling Tabular Data Manipulation by LLMs in Real Office Usage Scenarios},
  author={Xiaokang Zhang and Jing Zhang and Zeyao Ma and Yang Li and Bohan Zhang and Guanlin Li and Zijun Yao and Kangli Xu and Jinchang Zhou and Daniel Zhang-li and Jifan Yu and Shu Zhao and Juan-Zi Li and Jie Tang},
  journal={ArXiv},
  year={2024},
  volume={abs/2403.19318},
  url={https://api.semanticscholar.org/CorpusID:268732926}
}

@article{Yang2022CROPZC,
  title={CROP: Zero-shot Cross-lingual Named Entity Recognition with Multilingual Labeled Sequence Translation},
  author={Jian Yang and Jian Yang and Shaohan Huang and Shuming Ma and Yuwei Yin and Li Dong and Dongdong Zhang and Hongcheng Guo and Zhoujun Li and Furu Wei},
  journal={ArXiv},
  year={2022},
  volume={abs/2210.07022},
  url={https://api.semanticscholar.org/CorpusID:252873187}
}

@inproceedings{Yang2020AlternatingLM,
  title={Alternating Language Modeling for Cross-Lingual Pre-Training},
  author={Jian Yang and Shuming Ma and Dongdong Zhang and Shuangzhi Wu and Zhoujun Li and Ming Zhou},
  booktitle={AAAI Conference on Artificial Intelligence},
  year={2020},
  url={https://api.semanticscholar.org/CorpusID:214493073}
}

@article{Park2023MachineLB,
  title={Machine Learning based Post Event Analysis for Cybersecurity of Cyber-Physical System},
  author={Kuchan Park and Junho Hong and Wencong Su and HyoJong Lee},
  journal={ArXiv},
  year={2023},
  volume={abs/2311.13488},
  url={https://api.semanticscholar.org/CorpusID:265352134}
}

@article{Huang2022MBCTTF,
  title={MBCT: Tree-Based Feature-Aware Binning for Individual Uncertainty Calibration},
  author={Siguang Huang and Yunli Wang and Lili Mou and Huayue Zhang and Han Zhu and Chuan Yu and Bo Zheng},
  journal={Proceedings of the ACM Web Conference 2022},
  year={2022},
  url={https://api.semanticscholar.org/CorpusID:246680021}
}

@article{Zhai2024ActionsSL,
  title={Actions Speak Louder than Words: Trillion-Parameter Sequential Transducers for Generative Recommendations},
  author={Jiaqi Zhai and Lucy Liao and Xing Liu and Yueming Wang and Rui Li and Xuan Cao and Leon Gao and Zhaojie Gong and Fangda Gu and Michael He and Yin-Hua Lu and Yu Shi},
  journal={ArXiv},
  year={2024},
  volume={abs/2402.17152},
  url={https://api.semanticscholar.org/CorpusID:268033327}
}

@article{Jin2024ZeroShotCR,
  title={Zero-Shot Chain-of-Thought Reasoning Guided by Evolutionary Algorithms in Large Language Models},
  author={Feihu Jin and Yifan Liu and Ying Tan},
  journal={ArXiv},
  year={2024},
  volume={abs/2402.05376},
  url={https://api.semanticscholar.org/CorpusID:267547394}
}

@INPROCEEDINGS {10386611,
author = {J. Liu and J. Zhan},
booktitle = {2023 IEEE International Conference on Big Data (BigData)},
title = {Constructing Knowledge Graph from Cyber Threat Intelligence Using Large Language Model},
year = {2023},
volume = {},
issn = {},
pages = {516-521},
doi = {10.1109/BigData59044.2023.10386611},
url = {https://doi.ieeecomputersociety.org/10.1109/BigData59044.2023.10386611},
publisher = {IEEE Computer Society},
address = {Los Alamitos, CA, USA},
month = {dec}
}

@inproceedings{ganlm,
  author       = {Jian Yang and
                  Shuming Ma and
                  Li Dong and
                  Shaohan Huang and
                  Haoyang Huang and
                  Yuwei Yin and
                  Dongdong Zhang and
                  Liqun Yang and
                  Furu Wei and
                  Zhoujun Li},
  editor       = {Anna Rogers and
                  Jordan L. Boyd{-}Graber and
                  Naoaki Okazaki},
  title        = {GanLM: Encoder-Decoder Pre-training with an Auxiliary Discriminator},
  booktitle    = {Proceedings of the 61st Annual Meeting of the Association for Computational
                  Linguistics (Volume 1: Long Papers), {ACL} 2023, Toronto, Canada,
                  July 9-14, 2023},
  pages        = {9394--9412},
  publisher    = {Association for Computational Linguistics},
  year         = {2023},
  url          = {https://doi.org/10.18653/v1/2023.acl-long.522},
  doi          = {10.18653/V1/2023.ACL-LONG.522},
  bibsource    = {dblp computer science bibliography, https://dblp.org}
}

@inproceedings{paeg,
  author       = {Juncheng Wan and
                  Jian Yang and
                  Shuming Ma and
                  Dongdong Zhang and
                  Weinan Zhang and
                  Yong Yu and
                  Zhoujun Li},
  editor       = {Nicoletta Calzolari and
                  Chu{-}Ren Huang and
                  Hansaem Kim and
                  James Pustejovsky and
                  Leo Wanner and
                  Key{-}Sun Choi and
                  Pum{-}Mo Ryu and
                  Hsin{-}Hsi Chen and
                  Lucia Donatelli and
                  Heng Ji and
                  Sadao Kurohashi and
                  Patrizia Paggio and
                  Nianwen Xue and
                  Seokhwan Kim and
                  Younggyun Hahm and
                  Zhong He and
                  Tony Kyungil Lee and
                  Enrico Santus and
                  Francis Bond and
                  Seung{-}Hoon Na},
  title        = {{PAEG:} Phrase-level Adversarial Example Generation for Neural Machine
                  Translation},
  booktitle    = {Proceedings of the 29th International Conference on Computational
                  Linguistics, {COLING} 2022, Gyeongju, Republic of Korea, October 12-17,
                  2022},
  pages        = {5085--5097},
  publisher    = {International Committee on Computational Linguistics},
  year         = {2022},
  url          = {https://aclanthology.org/2022.coling-1.451},
  bibsource    = {dblp computer science bibliography, https://dblp.org}
}

@article{Lei2023CFBenchmarkCF,
  title={CFBenchmark: Chinese Financial Assistant Benchmark for Large Language Model},
  author={Yang Lei and Jiangtong Li and Ming Jiang and Junjie Hu and Dawei Cheng and Zhijun Ding and Changjun Jiang},
  journal={ArXiv},
  year={2023},
  volume={abs/2311.05812},
  url={https://api.semanticscholar.org/CorpusID:265128579}
}

@article{Li2023CFGPTCF,
  title={CFGPT: Chinese Financial Assistant with Large Language Model},
  author={Jiangtong Li and Yuxuan Bian and Guoxuan Wang and Yang Lei and Dawei Cheng and Zhijun Ding and Changjun Jiang},
  journal={ArXiv},
  year={2023},
  volume={abs/2309.10654},
  url={https://api.semanticscholar.org/CorpusID:262054189}
}

@article{Li2023TextbooksAA,
  title={Textbooks Are All You Need II: phi-1.5 technical report},
  author={Yuan-Fang Li and S{\'e}bastien Bubeck and Ronen Eldan and Allison Del Giorno and Suriya Gunasekar and Yin Tat Lee},
  journal={ArXiv},
  year={2023},
  volume={abs/2309.05463},
  url={https://api.semanticscholar.org/CorpusID:261696657}
}

@article{Narayan2018DontGM,
  title={Don’t Give Me the Details, Just the Summary! Topic-Aware Convolutional Neural Networks for Extreme Summarization},
  author={Shashi Narayan and Shay B. Cohen and Mirella Lapata},
  journal={ArXiv},
  year={2018},
  volume={abs/1808.08745},
  url={https://api.semanticscholar.org/CorpusID:215768182}
}

@article{Wei2022ChainOT,
  title={Chain of Thought Prompting Elicits Reasoning in Large Language Models},
  author={Jason Wei and Xuezhi Wang and Dale Schuurmans and Maarten Bosma and Ed Huai-hsin Chi and F. Xia and Quoc Le and Denny Zhou},
  journal={ArXiv},
  year={2022},
  volume={abs/2201.11903},
  url={https://api.semanticscholar.org/CorpusID:246411621}
}

@article{Young2024YiOF,
  title={Yi: Open Foundation Models by 01.AI},
  author={01.AI Alex Young and Bei Chen and Chao Li and Chengen Huang and Ge Zhang and Guanwei Zhang and Heng Li and Jiangcheng Zhu and Jianqun Chen and Jing Chang and Kaidong Yu and Peng Liu and Qiang Liu and Shawn Yue and Senbin Yang and Shiming Yang and Tao Yu and Wen Xie and Wenhao Huang and Xiaohui Hu and Xiaoyi Ren and Xinyao Niu and Pengcheng Nie and Yuchi Xu and Yudong Liu and Yue Wang and Yuxuan Cai and Zhenyu Gu and Zhiyuan Liu and Zonghong Dai},
  journal={ArXiv},
  year={2024},
  volume={abs/2403.04652},
  url={https://api.semanticscholar.org/CorpusID:268264158}
}

@article{Zeng2024ChatGLMAF,
  title={ChatGLM: A Family of Large Language Models from GLM-130B to GLM-4 All Tools},
  author={Team Glm Aohan Zeng and Bin Xu and Bowen Wang and Chenhui Zhang and Da Yin and Diego Rojas and Guanyu Feng and Hanlin Zhao and Hanyu Lai and Hao Yu and Hongning Wang and Jiadai Sun and Jiajie Zhang and Jiale Cheng and Jiayi Gui and Jie Tang and Jing Zhang and Juanzi Li and Lei Zhao and Lindong Wu and Lucen Zhong and Ming-yue Liu and Minlie Huang and Peng Zhang and Qinkai Zheng and Rui Lu and Shuaiqi Duan and Shudan Zhang and Shulin Cao and Shuxun Yang and Weng Lam Tam and Wenyi Zhao and Xiao Liu and Xiaoyu Xia and Xiaohan Zhang and Xiaotao Gu and Xin Lv and Xinghan Liu and Xinyi Liu and Xinyue Yang and Xixuan Song and Xunkai Zhang and Yi An and Yifan Xu and Yilin Niu and Yuantao Yang and Yueyan Li and Yushi Bai and Yuxiao Dong and Zehan Qi and Zhaoyu Wang and Zhenyi Yang and Zhengxiao Du and Zhen-Ping Hou and Zihan Wang},
  journal={ArXiv},
  year={2024},
  volume={abs/2406.12793},
  url={https://api.semanticscholar.org/CorpusID:270562306}
}

@inproceedings{Yang2024Qwen2TR,
  title={Qwen2 Technical Report},
  author={An Yang and Baosong Yang and Binyuan Hui and Bo Zheng and Bowen Yu and Chang Zhou and Chengpeng Li and Chengyuan Li and Dayiheng Liu and Fei Huang and Guanting Dong and Haoran Wei and Huan Lin and Jialong Tang and Jialin Wang and Jian Yang and Jianhong Tu and Jianwei Zhang and Jianxin Ma and Jin Xu and Jingren Zhou and Jinze Bai and Jinzheng He and Junyang Lin and Kai Dang and Keming Lu and Ke-Yang Chen and Kexin Yang and Mei Li and Min Xue and Na Ni and Pei Zhang and Peng Wang and Ru Peng and Rui Men and Ruize Gao and Runji Lin and Shijie Wang and Shuai Bai and Sinan Tan and Tianhang Zhu and Tianhao Li and Tianyu Liu and Wenbin Ge and Xiaodong Deng and Xiaohuan Zhou and Xingzhang Ren and Xinyu Zhang and Xipin Wei and Xuancheng Ren and Yang Fan and Yang Yao and Yichang Zhang and Yunyang Wan and Yunfei Chu and Zeyu Cui and Zhenru Zhang and Zhi-Wei Fan},
  year={2024},
  url={https://api.semanticscholar.org/CorpusID:271212307}
}

@article{Araci2019FinBERTFS,
  title={FinBERT: Financial Sentiment Analysis with Pre-trained Language Models},
  author={Dogu Araci},
  journal={ArXiv},
  year={2019},
  volume={abs/1908.10063},
  url={https://api.semanticscholar.org/CorpusID:201646244}
}

@article{Cui2023EfficientAE,
  title={Efficient and Effective Text Encoding for Chinese LLaMA and Alpaca},
  author={Yiming Cui and Ziqing Yang and Xin Yao},
  journal={ArXiv},
  year={2023},
  volume={abs/2304.08177},
  url={https://api.semanticscholar.org/CorpusID:258180548}
}

\end{document}